\documentclass[a4paper,conference]{IEEEtran}
\usepackage{algorithm}
\usepackage{algorithmic}

\usepackage{amsmath}
\usepackage{amssymb}
\usepackage{times,graphicx,tabulary,multirow,xspace,xcolor}
\usepackage[inline, shortlabels]{enumitem}
\usepackage{color, colortbl}
\definecolor{LightGray}{rgb}{0.95,0.95,0.95}

\newcommand{\jing}[1]{{\color[rgb]{1,0,0}{\textbf{JH:}}{\normalsize\itshape#1}}}

\usepackage{pifont}
\newcommand{\cmark}{\color{gray}\ding{51}}%
\newcommand{\xmark}{\color{gray}\ding{55}}%

\newcommand{\belowsec}{0pt}

\newcommand{\eat}[1]{}

\def\eg{\emph{e.g}.} 
\def\ie{\emph{i.e}.} 
 
\def\etc{\emph{etc}.} 
 
\def\etal{\emph{et al}.}

\usepackage[pagebackref=true,breaklinks=true,colorlinks=false,bookmarks=false]{hyperref}

\title{Cross-modal Contrastive Distillation for Instructional Activity Anticipation}

\author{\IEEEauthorblockN{Zhengyuan Yang$^1$, Jingen Liu$^2$, Jing Huang$^2$, Xiaodong He$^2$, Tao Mei$^2$, Chenliang Xu$^1$, Jiebo Luo$^1$}
\IEEEauthorblockA{$^1$Department of Computer Science, University of Rochester \qquad $^2$JD AI Research}
}
\begin{document}

\maketitle

\begin{abstract}
  In this study, we aim to predict the plausible future action steps given an observation of the past and study the task of instructional activity anticipation. Unlike previous anticipation tasks that aim at action label prediction, our work targets at generating natural language outputs that provide interpretable and accurate descriptions of future action steps. It is a challenging task due to the lack of semantic information extracted from the instructional videos. To overcome this challenge, we propose a novel knowledge distillation framework to exploit the related external textual knowledge to assist the visual anticipation task. However, previous knowledge distillation techniques generally transfer information within the same modality. To bridge the gap between the visual and text modalities during the distillation process, we devise a novel cross-modal contrastive distillation (CCD) scheme, which facilitates knowledge distillation between teacher and student in heterogeneous modalities with the proposed cross-modal distillation loss. We evaluate our method on the Tasty Videos dataset. CCD improves the anticipation performance of the visual-alone student model by a large margin of $40.2\%$ relatively in BLEU4. Our approach also outperforms the state-of-the-art approaches by a large margin.
\end{abstract}

\section{Introduction}

Video activity anticipation aims to predict the plausible future action steps based on partial observation of a video. Activity anticipation can be used for various interesting applications such as robotics, motion planning, and general video understanding. Previous anticipation studies explore various video types, including regular videos with human actions~\cite{ryoo2011human,xu2015activity,vondrick2016anticipating}, instructional videos~\cite{sener2019zero,abu2018will,ke2019time,chang2020procedure}, or videos for specific scenes such as vehicle driving~\cite{chang2019argoverse,chandra2019traphic} or pedestrian walking~\cite{alahi2016social,ma2017forecasting}. Based on the output format, the predicted actions can be represented as action labels~\cite{ryoo2011human,hoai2014max,xu2015activity,lan2014hierarchical,abu2018will,ke2019time}, visual representations~\cite{vondrick2016anticipating,rodriguez2018action}, or natural language descriptions~\cite{sener2019zero}. Unlike the mainstream studies that anticipate action labels~\cite{ryoo2011human,hoai2014max,xu2015activity,lan2014hierarchical,abu2018will,ke2019time}, we instead focus on generating language captions on instructional videos that provide an interpretable and accurate description of the anticipated future activities. 

In contrast to the high processing cost of instructional videos, procedural knowledge is widely available in text descriptions, \eg, cooking recipes. Text description often provides a more precise description of the procedural action evolution and contains semantic textual information that requires less computation. In addition, motivated by how humans anticipate the future from their knowledge and experiences, we aim to assist video activity anticipation with related textual knowledge. Therefore in this paper, we present a new framework that utilizes textual knowledge to assist video anticipation. Our framework can also be used to integrate other forms of knowledge, such as video scene graph~\cite{ji2020action}, in addition to textual knowledge.

Integrating related knowledge from different modalities for activity anticipation is nontrivial. A previous study~\cite{sener2019zero} approached this problem by reusing a pre-trained text model's weights for the visual task. The method makes use of the text knowledge by strictly fitting a visual encoder from a pre-trained text anticipation model. Despite its success in some scenarios, such a heuristic weight reusing approach requires the visual model to have an identical network architecture as the textual model, making it less flexible to apply. Furthermore, reusing the weights assumes that a good anticipation module (trained on the text) works equally well for both the visual and text inputs. However, this assumption does not hold as previous study~\cite{sener2019zero} observes a performance drop when reusing the text anticipation module for visual anticipation. Intuitively, it would be beneficial to fine-tune the anticipation module with visual information while enforcing the model to remember knowledge from the pre-trained text model.

To this end, we propose a new framework that is flexible and effective to assist video anticipation with textual knowledge. The core idea is to use knowledge distillation to transfer the procedural information from text modality to visual modality, where we consider the text model as the teacher and the video model as the student. Knowledge distillation~\cite{hinton2015distilling,romero2014fitnets,zagoruyko2016paying,gupta2016cross,li2017learning} has shown its effectiveness in various tasks by providing better supervision signals. The main idea is to extract information from the teacher model, \eg, prediction logits~\cite{hinton2015distilling}, intermediate features~\cite{romero2014fitnets}, or model weights, and use them as additional supervision signals to train student models. However, the teacher and student in conventional knowledge distillation methods take the inputs from the same modality to form paired supervision signals. Thus, previous distillation methods cannot be directly applied to video anticipation with textual knowledge, as the teacher (text) and student (video) take inputs in different modalities.

To solve the above problem, we propose cross-modal contrastive distillation (CCD),\eat{ as shown in Figure~\ref{fig:intro},} which allows the teacher and student to take inputs from different modalities with a new cross-modal contrastive loss. In CCD, the inputs to the teacher and student models are pairs with the same semantic meaning but in different modalities. For example, a cooking video containing multiple steps may have a corresponding multi-step text recipe instruction. CCD conducts distillation at the feature level. Instead of directly minimizing the absolute distance in feature space between the teacher-student pair, CCD adopts a novel contrastive loss that only requires preserving the relative similarity, \ie, the pairwise relationship between teacher-student pairs. With the proposed cross-modal contrastive loss for distillation, we successfully handle the cross-modal distillation problem and transfer the semantically relevant information in the text modality to assist video anticipation. Because text is less computationally demanding than video, we can acquire stronger teacher models by utilizing the abundant external procedural text data. CCD then efficiently transfers the text knowledge to the student visual anticipation model, facilitating the use of extra prior knowledge with a minimal number of required video training samples.

In addition, our single-modal visual model (student) and textual model (teacher) in the distillation framework are built with the transformer units. The multi-head attention mechanism enables our single-modal model to handle long sequences more effectively than the RNN-based models~\cite{venugopalan2015sequence,sener2019zero}\eat{ than RNN in~\cite{sener2019zero} \jing{add comparison here, more citations with RNN model please}}, thus boosting the anticipation performance. We thoroughly benchmark our method on the Tasty Videos dataset~\cite{sener2019zero} with our proposed transformer-based anticipation network. By effectively utilizing the procedural knowledge in text via the proposed CCD, our method significantly outperforms the state of the art and the visual-alone baseline. Our method improves the state-of-the-art BLEU4 score~\cite{sener2019zero} from $1.48$ to $7.36$. Compared with the visual-only baseline, CCD obtains a $2.1$ absolute gain in BLEU4, which amounts to a $40.2\%$ relative improvement.

In summary, our main contributions are:
\begin{itemize} 
\item We propose cross-modal contrastive distillation (CCD) that facilities distilling teacher's information to the student in a different modality. To the best of our knowledge, CCD is the first method that introduces knowledge distillation modeling for video anticipation.
\item We apply CCD to the video activity anticipation task and use the procedural information in the text to help video activity anticipation.
\item We benchmark CCD with our proposed transformer-based network. Our method achieves the new state of the art on the Tasty Videos Dataset.
\vspace{-3pt}
\end{itemize}
\section{Related Work}
\noindent\textbf{Video anticipation.} Various approaches~\cite{abu2018will,furnari2019would,farha2020long,ng2020forecasting} have been developed for the video anticipation task.
A common anticipation pipeline is based on a sequence-to-sequence model. The observed video clips are first encoded and inputted to the temporal modeling module as a set of feature vectors. The temporal modeling module anticipates the future steps as a set of feature vectors. A decoding module then decodes the predicted vectors for future steps into the anticipation results, \eg, action labels, text descriptions, or visual representations.

Most previous video anticipation studies are intuitively based on the single visual modality. However, the activity procedure knowledge could exist in other modalities, such as text and action graph~\cite{ji2020action}. This observation motivates our study to use the procedure knowledge in another modality to help activity anticipation. Our study is most related to \cite{sener2019zero} where the authors propose to utilize the text recipe knowledge to help the cooking activity anticipation task. Specifically, the model fixes and reuses the text model's parameters by fitting a visual encoder to the remainder of the pre-trained model. The major difference is that we explore effective and flexible framework of knowledge transfer based on knowledge distillation from two different modalities, in contrast to reusing model weights as in the previous study~\cite{sener2019zero}.

\begin{figure*}[t]
\begin{center}
  \centerline{\includegraphics[width=14cm]{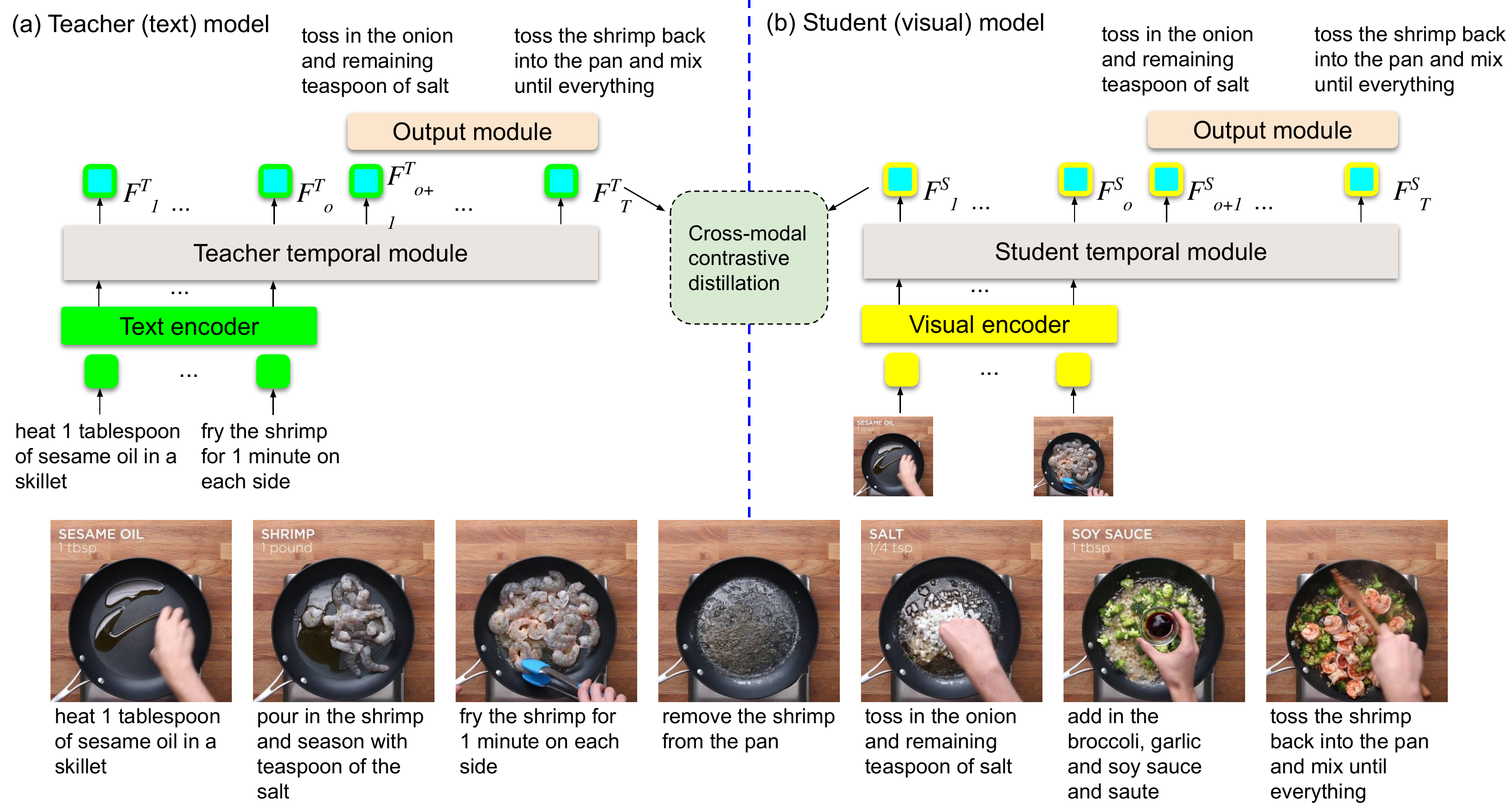}}
\end{center}
\vspace{-0.3in}
    \caption{The proposed cross-modal contrastive distillation (CCD) framework with the transformer models for instructional activity anticipation. CCD distills the text information (teacher) to the main visual anticipation task (student).
    By enforcing the similarity of the corresponding intermediate features, such as $F^T$ and $F^S$, CCD transfers the procedural knowledge in text to help visual activity anticipation.
	}
\vspace{-0.2in}
\label{fig:arch}
\end{figure*}

\noindent\textbf{Knowledge distillation.}
Our proposed CCD is built based on the knowledge distillation technique. Knowledge distillation~\cite{hinton2015distilling,romero2014fitnets,zagoruyko2016paying,gupta2016cross,li2017learning} trains a small student model by mimicking the output distribution~\cite{hinton2015distilling}, activation of intermediate layers~\cite{romero2014fitnets}, or attention maps~\cite{zagoruyko2016paying} of a large teacher model. Our proposed contrastive distillation method is related to \cite{tian2019contrastive} that also leverages contrastive learning for distillation. The major difference is that we extend the distillation between highly-paired and homogeneous modalities of RGB and depth images~\cite{tian2019contrastive} to the heterogeneous modalities of texts and videos. Knowledge distillation has also shown effectiveness in various vision-language tasks, such as VQA~\cite{mun2018learning,do2019compact,fang2021compressing}, grounded image captioning~\cite{zhou2020more}, video caption~\cite{pan2020spatio,zhang2020object}, \etc.

Compared with the conventional knowledge distillation methods, our proposed CCD considers a special case where the teacher and student are in different modalities. CCD adopts contrastive learning~\cite{karpathy2015deep,faghri2017vse++,li2019visual} as the distillation loss function. We show in experiments that the conventional distillation methods fail in transferring the procedure knowledge, while the proposed CCD effectively improves activity anticipation performance with the distilled text knowledge.

\noindent\textbf{Text knowledge for vision tasks.}
Recent studies~\cite{tsimpoukelli2021multimodal,yang2021empirical} explore using large-scale language models~\cite{brown2020language} for vision tasks and observe significant improvements with text knowledge. We expect CCD provides a better way of utilizing the text knowledge, thus further improving the vision task performance.
\section{Approach}
In this section, we first introduce the problem definition and our proposed network for the video activity anticipation task. We then describe our proposed CCD framework.

\subsection{Problem definition}
\label{sec:input}
The input to the video activity anticipation task is the partial observation of a video, and the expected output is the anticipated instructions in future steps. We assume the long video consists of $T$ steps, and the input to the system are the video clips from $1$ to $o$, where $o\in \{1,\cdots,T-1 \}$. The output of the system is the predicted instructions for each future step as a natural language sentence. Specifically, for each step $s_{o+1}$ to $s_T$, we generate a text output for the predicted instruction. Note that we take the ingredients keywords as the input (effectively as the $0$-th step $s_0$).

\subsection{Transformer model for activity anticipation}
\label{sec:trans}
We propose to use the transformer model as the activity anticipation model as shown in Figure~\ref{fig:arch}. We design the network in a sequence-to-sequence architecture and use transformer ~\cite{vaswani2017attention} as the model's base units. The sequence-to-sequence model consists of three major components, \ie, the input text/visual encoder, temporal module for anticipating future activity in the hidden space, and the output module that converts the anticipated future into the textual caption. In the following we introduce the framework and the anticipation task modeling in details.

\noindent\textbf{Input visual encoder.}
We encode each observed input video clip by max-pooling over the extracted ResNet-50~\cite{he2016deep} frame features, and then project it into a $d$-dimension vector with linear transforms. We represent the ingredients as an N-hot vector and learn an embedding matrix that projects the N-hot vector into an ingredient feature. We then project the embedded feature into a $d$-dimension vector with linear transforms.

\noindent\textbf{Temporal module.}
The temporal module in the instruction anticipation transformer aims to anticipate future instructions given the partial observations. The module consists of $M$ layers of transformer~\cite{vaswani2017attention}. We adopt the causal attention masks~\cite{zhou2020unified,li2020oscar,yang2021tap} to make the transformer only sees the previous steps. The input to the temporal module is the $d$-dimension encoded features for video clips and text ingredients in previous steps. The output feature is $T$ anticipated features $F_t, t\in \{1,\cdots,T \}$ at all time steps.

\noindent\textbf{Output module.}
At each future time-step $t \in \{o+1,\cdots,T \}$, the output module is applied independently and converts the anticipated features $F_t$ into language instruction. The output module consists of $N$ transformer layers with causal attention masks. The output modules in different time-steps share weights. The entire sequence-to-sequence model is trained end-to-end with the anticipated captions in future steps $t \in \{o+1,\cdots,T \}$. In each step, the model is trained with teacher-forcing~\cite{lamb2016professor}\eat{, \ie, using ground-truth inputs in the output module}. The computed caption losses in all steps are combined and jointly optimize the entire model.

\subsection{Cross-modal contrastive distillation}
\label{sec:ccd}
In this study, we explore using the procedural information in the text modality (e.g. text recipe) to help the visual anticipation task via knowledge distillation. Despite the same semantic procedural information, the feature in teacher and student are in different modalities, which fails the conventional knowledge distillation methods. Therefore we propose cross-modal contrastive distillation (CCD) to transfer knowledge among different modalities.

\noindent\textbf{Teacher-student models.}
We first formally introduce the student and teacher settings. We consider the text modality as the teacher. The input to the teacher model is a similar partial observation of a sequence as in the problem definition section, except the observation in each clip is a sentence instead of a video clip. We encoder the observed sentence in each time step with a transformer encoder respectively, and map the encoded feature with linear transforms into a $d$-dimensional feature. We train a separate ``instruction anticipation transformer'' with the same transformer architecture for the text teacher model. Therefore, the teacher model is a textual sequence-to-sequence anticipation model, while the visual anticipation transformer is the student model.

From the teacher and student setting, we find correspondence between each step and intermediate features in both the temporal module and the output module. Intuitively, the constraint that enforces the corresponding features $F^S$ in the student model to be similar to the ones $F^T$ in the teacher model would transfer the text procedural information to the student and thus benefit the video anticipation task.

\noindent\textbf{Cross-modal contrastive distillation.}
Since $F^S$ and $F^T$ contain information in different modalities (text and visual), directly enforcing their similarity in feature space does not work well. Instead, we propose the cross-modal contrastive distillation loss to address this problem. Specifically, we apply the distillation loss in the form of triplet loss~\cite{karpathy2015deep,faghri2017vse++,li2019visual} that keeps the correspondence between the features in the teacher and student models:
\begin{align*}
        \mathcal{L}_{ccd} = \sum_{k=1}^K &\left\{ \left[ \alpha- S(F_k^S,F_k^T) + S(F_k^S,F_i^T) \right ]_+ \right . \\
+ & \left . \left[ \alpha- S(F_k^S,F_k^T) + S(F_j^S,F_k^T) \right ]_+ \right\},
\end{align*}
where $\alpha$ is the margin with a default value of $0.2$ and can be tuned on a validation set. $\text{S}(\cdot)$ is the similarity function in the feature space. We use inner project with a learnable linear transform over $F^{S}$ and $F^{T}$ as $\text{S}(\cdot)$ in our experiments. $i,j$ are the index for the hard negatives where $i=\text{argmax}_{i\neq k} S(F_k^{S},F_{i}^{T})$ and $j=\text{argmax}_{j\neq k} S(F_{j}^{S},F_k^{T})$. $k$ is the index for the selected feature for distillation. We detail the selected features and their combinations in the experimental section.

With the proposed contrastive distillation loss, CCD only enforces the relative similarity between the student-teacher feature pairs, instead of conventional distillation objectives (\eg, KL divergence or L2 loss) that tries to make the student-teacher features 
close in feature space. We find this relative similarity important in the cross-modal distillation scenarios as in our text-assisted video anticipation task. Despite the same semantic meaning, teacher and student are inherently in different modalities. Because of this, enforcing student-teacher's features to be identical does not benefit the student model, as detailed in experiment results. 

We train the entire model end-to-end with the combination of the teacher-forcing caption loss $\mathcal{L}_{cap}$ and the contrastive knowledge distillation loss $\mathcal{L}_{ccd}$:
\begin{equation}
\label{eq:loss}
    \mathcal{L} = \mathcal{L}_{cap} +\mathcal{L}_{ccd} * w_{ccd},
\end{equation}
where $w_{ccd}$ is the distillation loss weight with a default value of $10$ and can be tuned on a validation set. Here we compute a single CCD loss $\mathcal{L}_{ccd}$ over $F^S$ and $F^T$ for simplicity. In practice, we can apply CCD loss to different student-teacher feature pairs, for example, the input to the temporal module, the temporal module's output, and the transformer's hidden states. Detailed ablation studies are in the ablation section.
\section{Experiments}

\subsection{Datasets}
\label{sec:dataset}
\noindent\textbf{Tasty video dataset.}
The Tasty Videos dataset~\cite{sener2019zero} has 2511 cooking instructional videos with paired multi-step text recipes. Each video contains multi-step annotation with the video clip and per-step text instruction. Each video also includes a list of ingredients. We follow the training/validation/testing split in previous studies~\cite{sener2019zero}. The average length of the video is 1551 frames or 54 seconds. There are nine steps on average per video.

\noindent\textbf{Recipe1M dataset.}
We use the Recipe1M dataset~\cite{salvador2017learning} to strengthen the teacher model. Recipe1M contains around one million cooking instructions. Each sample contains multi-step text recipes, a paired image, and an ingredient list. We use Recipe1M's text information (recipe and ingredient) to boost the teacher model's performance. 

\begin{table}
\centering
\caption{Next-step anticipation performance on the Tasty Videos dataset. The upper part shows the text-based anticipation results where the input is recipe text, which serve as the text teacher model. The bottom part shows the main task of visual-based anticipation where the input is video clips. Rows (5-7) are the state-of-the-art methods and Rows (8-12) are the variants of our proposed method. We highlight the visual-alone baseline (Row 8) and our CCD model (Row 12) in {\color{gray}gray}. Row 4 shows the performance of the text teacher model.}
\vspace{-0.0in}
\begin{tabular}{c | l l c c}
    \hline
    \# & Method & Modality\eat{ & Backbone} & BLEU1 & BLEU4 \\
    \hline
    1 & Sener~\etal~\cite{sener2019zero} & Text\eat{ & RNN} & 10.58 & 0.24\\
    2 & Sener~\etal~\cite{sener2019zero}+Recipe1M & Text\eat{ & RNN} & 26.78 & 3.30 \\
    3 & Text-alone & Text\eat{ & Transformer} & 25.59 & 4.95 \\
    4 & Text-alone+Recipe1M & Text\eat{ & Transformer} & 32.88 & 11.77 \\
    \hline
    5 & S2VT~\cite{venugopalan2015sequence} & Visual\eat{ & RNN} & 9.14 & 0.26\\
    6 & Sener~\etal~\cite{sener2019zero} & Visual\eat{ & RNN} & 19.05 & 1.48\\
    7 & End-to-end~\cite{zhou2018end} & Visual\eat{ & Transformer} & - & 0.54 \\
    \rowcolor{LightGray}
    8 & Visual-alone & Visual\eat{ & Transformer} & 26.08 & 5.25 \\
    9 & Logits distillation & Visual\eat{ & Transformer} & 26.64 & 5.87 \\
    10 & Feature distillation & Visual\eat{ & Transformer} & 26.58 & 5.69 \\
    11 & Discriminator distillation & Visual\eat{ & Transformer} & 26.11 & 4.03 \\
    \rowcolor{LightGray}
    12 & CCD (Ours) & Visual\eat{ & Transformer} & \textbf{27.37} & \textbf{7.36} \\
    \hline
\end{tabular}
\vspace{-0.1in}
\label{table:main}
\end{table}
\subsection{Experimental settings}
\noindent\textbf{Evaluation metric.}
We focus on the next step anticipation problem in this study. We evaluate the model with the captioning metric BLEU~\cite{papineni2002bleu}.

\noindent\textbf{Implementation details.}
We use the vocabulary of $30,171$ from Recipe1M~\cite{salvador2017learning}. We find using the Recipe1M's food-related vocabulary important for getting good performance. We set the transformer layer numbers $M$ and $N$ to be $2$ and $3$, respectively. We train the transformer layers from scratch. The dimension $d$ of the instruction anticipation transformer is $768$. The max output length of the generated sentence is $24$. 

We train the model with the Adam~\cite{kingma2014adam} optimizer at a learning rate of $10^{-4}$. For the teacher model, we first train the text anticipation model with the extra data in Recipe1M~\cite{salvador2017learning} for $20$ epochs to strengthen the model. We then fine-tune the model on the Tasty Videos dataset with the text anticipation task for $40$ epochs. For student model's video anticipation task, we train the model for $40$ epochs with a batch size of $16$.

\noindent\textbf{Compared methods.}
In Table~\ref{table:main}, we compare CCD with the state of the art~\cite{sener2019zero} and the following baselines and alternatives.
\vspace{1pt}
\noindent\textit{Row 8: Visual-alone baseline.} We train visual-alone baseline from scratch based on the transformer anticipation model. The model's input is partial video clips without text information.

\vspace{1pt}
\noindent\textit{Row 9: Logits distillation.}
As an alternative to our proposed CCD, we keep the distillation modeling and use `logits distillation'': \ie, we add distillation losses over the instruction anticipations' predicted logits, in the form of KL divergence~\cite{hinton2015distilling}.

\vspace{1pt}
\noindent\textit{Row 10: Feature-distillation.} Distillation over the activation of intermediate layers is another common approach~\cite{romero2014fitnets}. We enforce the similarity between the paired activation in the teacher and student model with the L2 loss.

\vspace{1pt}
\noindent\textit{Row 11: Discriminator distillation.} Recent knowledge distillation studies~\cite{shen2020meal} show the effectiveness of boosting distillation performance with a teacher or student discriminator. In ``discriminator distillation,'' we experiment with adding a discriminator over either the final logits.

\begin{figure}[t]
\begin{center}
  \centerline{\includegraphics[width=8cm]{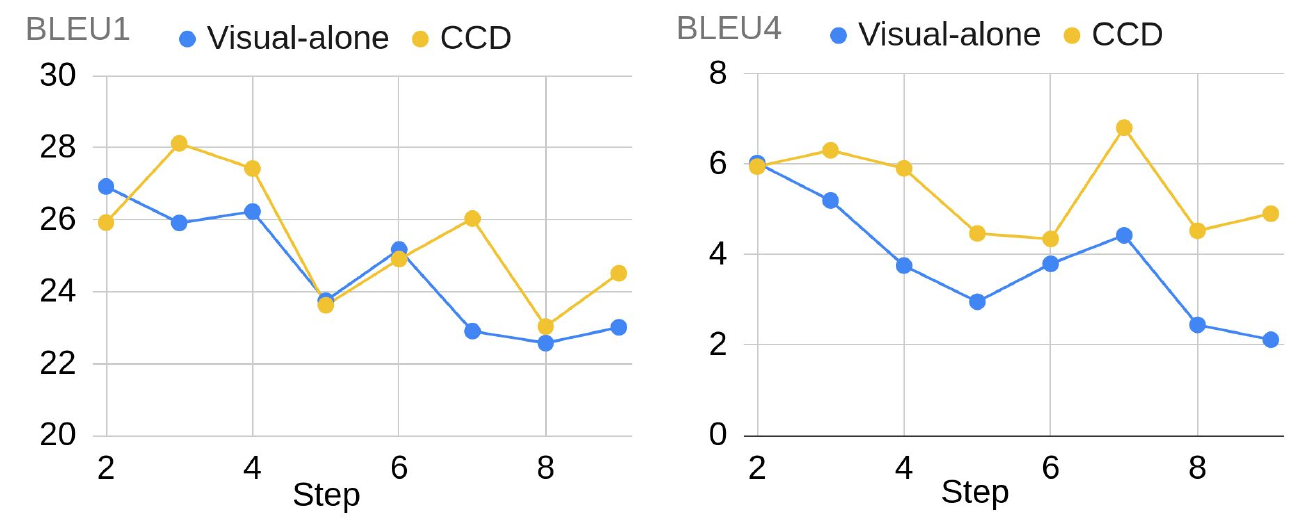}}
\end{center}
\vspace{-0.25in}
    \caption{Next-step predictions results on the Tasty Videos dataset. The x-axis shows anticipation steps and the y-axis shows BLEU scores. CCD consistently outperforms the visual-alone model in all steps.
	}
\vspace{-0.1in}
\label{fig:nextstep}
\end{figure}

\subsection{Instruction anticipation results}
\label{sec:result}
\vspace{\belowsec}
Table~\ref{table:main} shows the next step anticipation results on the Tasty Videos dataset. The upper part of the table shows text anticipation results, where the model input is the recipe text. The bottom part shows the main visual anticipation task.

\noindent\textbf{Single-modality evaluation.}
We first compare the methods that use the information from a single modality. Among text anticipation models (upper part), our transformer-based model performs better than previous RNN-based approaches (Text-alone: $4.95$, Sener~\etal~\cite{sener2019zero}: $0.24$). The motivation of our study is to exploit the large scale text knowledge to help video anticipation. Therefore, we utilize the large scale text data from the Recipe1M dataset to boost the text model. Both the RNN- and our transformer-based model improve significantly with the extra text data. Our text-alone method achieves the BLEU4 score of $11.77$, compared to the previous $4.95$. Meanwhile, our transformer-based model still outperforms RNN methods by a large margin of $8.47$ in BLEU4 (Rows 2,4).

Row 8 shows the anticipation performance of the visual-alone model. Our visual-alone model achieves a BLEU4 score of $5.25$, which is comparable to the text-alone performance of $4.95$. However with the extra Recipe1M data, the text model's performance improves to $11.77$, providing the possibility of serving as the teacher to help the visual model.

\noindent\textbf{CCD compared with the visual-alone model.}
We adopt the text model in Row 4 as the teacher model and use it to improve the visual-alone student model in Row 8 with our proposed CCD. As shown in Row 12, CCD outperforms the visual-alone model and the state of the art video anticipation methods~\cite{venugopalan2015sequence,zhou2018end,sener2019zero} (Rows 5-7) by large margins. CCD achieves a BLEU4 score of $7.36$ as in Row 12, compared with the visual-alone score of $5.25$ in Row 8.

We also compare the per-step performance of CCD and the visual-alone baseline in Figure~\ref{fig:nextstep}. We observe that CCD consistently outperforms the visual-alone model by effectively distilling the text procedural information. We observe larger improvements in later steps, \eg, steps $7,8,9$ in BLEU4. We conjecture that later steps rely more on the procedural information, while early steps have more fixed patterns such as ``mix'' for bakery, ``heat pan'' for stir-frying dishes, \etc. We present detailed analysis in qualitative results.

\noindent\textbf{Compared with other knowledge distillation methods.}
We also compare our proposed CCD to other conventional knowledge distillation methods, where we apply the same set of teacher and student models (Rows 4 and 8).
In the multi-modal setting, our proposed CCD performs better than conventional knowledge distillation methods. As shown in Rows 9-11, distillation over logits, intermediate features, or with discriminators shows limited improvements from the visual-alone baseline (Row 9: $5.87$, Row 8: $5.25$). In contrast, CCD improves the BLEU4 to $7.36$, as shown in Row 12. Although the teacher and student contain relevant procedural information, the features are in different modalities (text and visual). Therefore, enforcing the absolute similarity between teacher-student pair (by KL divergence or L2 loss) is too restrictive and thus hurts the main anticipation performance. Instead, our proposed CCD only requires preserving the teacher-student correspondence and thus performs better in the multi-modal setting.

\begin{table}
\centering
\caption{Ablation studies on the CCD loss position.}
\vspace{-0.1in}
\begin{tabular}{c | c c c c | c c c}
    \hline
    & Clip & Dec. & Temp. & Output  & BLEU1 & BLEU4 \\
    \hline
    (a) & - & - & - & - & 26.08 & 5.25 \\
    \rowcolor{LightGray}
    (b) & \cmark & - & - & - & 27.15 & 7.13 \\
    \rowcolor{LightGray}
    (c) & - & \cmark & - & - & 27.80 & 7.27 \\
    \rowcolor{LightGray}
    (d) & - & - & \cmark & - & 25.80 & 6.11 \\
    \rowcolor{LightGray}
    (e) & - & - & - & \cmark & 26.36 & 6.27 \\
    (f) & \cmark & \cmark & - & - & 27.12 & 7.23 \\
    (g) & - & - & \cmark & \cmark & 26.42 & 6.33 \\
    \rowcolor{LightGray}
    (h) & \cmark & \cmark & \cmark & \cmark & \textbf{27.37} & \textbf{7.36} \\
    \hline
\end{tabular}
\vspace{-0.1in}
\label{table:pos}
\end{table}
\begin{table}
\centering
\caption{Ablation studies on the teacher and student models that have different network architectures. CCD can flexibly transfers the teacher knowledge to students with different architectures, and improves the student performance.}
\vspace{-0.1in}
\begin{tabular}{c | c c c c}
    \hline
    & Student dim. & CCD & BLEU1 & BLEU4 \\
    \hline
    (a) & 768 & \xmark & 26.08 & 5.25 \\
    (b) & 768 & \cmark & 27.37 & 7.36 \\
    \hline
    (c) & 192 & \xmark & 27.63 & 6.51 \\
    (d) & 192 & \cmark & 27.67 & 6.85 \\
    \hline
\end{tabular}
\vspace{-0.1in}
\label{table:192}
\end{table}
\begin{figure*}[t]
\begin{center}
  \centerline{\includegraphics[width=16.6cm]{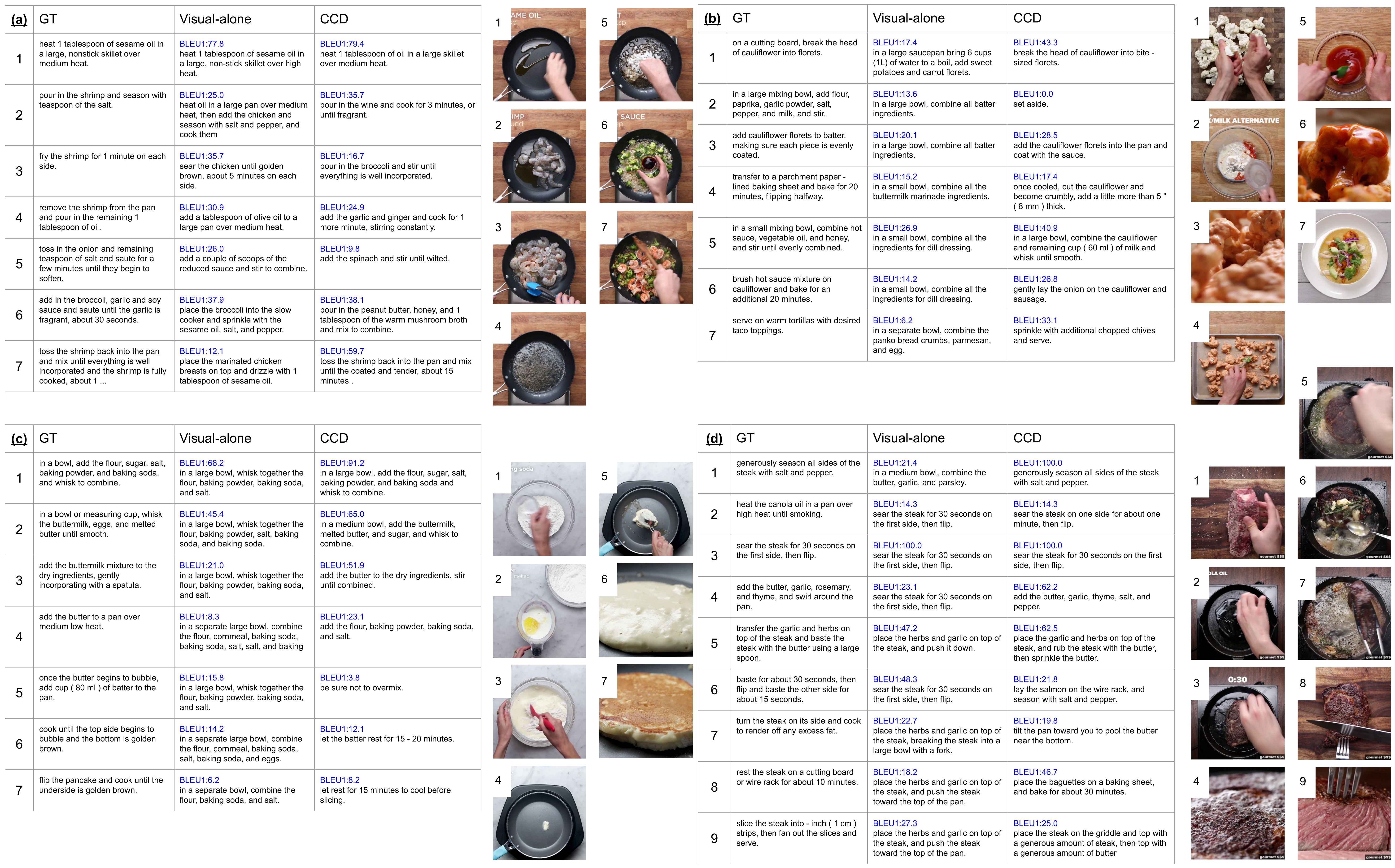}}
\end{center}
\vspace{-0.3in}
    \caption{Representative qualitative examples of visual-alone and CCD anticipation results. Best viewed on screen by zoom-in.
	}
\vspace{-0.2in}
\label{fig:visu_full}
\end{figure*}
\subsection{Ablation studies}
\label{sec:ablation}
\vspace{\belowsec}
\noindent\textbf{CCD loss position.}
We experiment with adding the CCD loss over different features between the teacher and student. Specifically, we experiment with the four features, \ie, the clip feature, decoding feature, temporal module's hidden state, and output module's hidden state. The clip feature is the $768$-dimensional encoded text or visual feature input to the temporal module. The decoding feature is the $768$-dimensional output feature $F^S$ or $F^T$ output from the temporal module. We also experiment with the $M\times 768$ and $N\times 768$ hidden states in the temporal and output module. We apply the CCD loss between each layer's hidden states.

Table~\ref{table:pos} explores where to apply the CCD loss, \ie, on the clip features (Clip), decoding features (Dec.), temporal hidden states (Temp.), or output hidden states (Output). Row $(a)$ is the visual-alone baseline without external text knowledge. We observe consistent improvements of CCD's variants over the visual-alone model (rows $(b$-$h)$). When applied independently, the CCD loss over the decoding feature shows the best performance ($7.27$ in row $(c)$). Jointly using losses over all features further improves the performance ($7.36$ in row $(h)$).

\noindent\textbf{Different teacher-student architectures.}
One advantage of transferring text procedural information with CCD is that CCD does not require the network architecture of the teacher and student to be identical. Therefore, the text procedural knowledge distillation can be more flexibly applied among different teacher-student models. We take a smaller student model with a hidden dimension of $192$ instead of $768$ in previous experiments. We add a linear projection in the CCD loss to match the feature dimension sizes of the teacher and student models. Table~\ref{table:192} shows the cross-modal distillation results on the smaller student model. The improvement of rows $(d)$ over rows $(c)$ indicates that CCD can be flexibly applied to transfer the teacher knowledge into the student model with different network architectures. Compared with row $(b)$, the improvement of using CCD is smaller on the small student model in row $(d)$. Compared with row $(a)$, the smaller student model in row $(c)$ shows a slightly better performance. We conjecture that a smaller model's better performance is caused by the limited training data in the Tasty Videos dataset, where a smaller model can prevent over-fitting. 

\subsection{Qualitative results}
\label{sec:quali}
Figure~\ref{fig:visu_full} visualizes anticipation results from visual-alone and CCD models. We observe CCD's following advantages.

\noindent\textit{(1) CCD generates predictions with better temporal evolution.}
Specifically, the visual-alone model frequently generates repeated or less meaningful anticipations, \eg, steps $2$-$7$ in Figure~\ref{fig:visu_full}(b) that repeats ``in a bowl, combine all ingredients.'' Similar predictions with poor temporal structure are observed in other examples, \eg, steps $1$-$7$ in Figure~\ref{fig:visu_full}(c) and steps $2$-$4$ in Figure~\ref{fig:visu_full}(d). In contrast, CCD generates better temporal anticipation by distilling the temporal procedural knowledge.

\noindent\textit{(2) CCD generates more semantically relevant predictions that are consistent with human perception.} In Figure~\ref{fig:visu_full}(a) step $3$, the visual-alone model predicts ``sear the chicken until golden brown, about 5 minutes on each side,'' and CCD predicts ``pour in the broccoli and stir until everything is well incorporated.'' Despite the visual-alone model's much higher BLEU score (visual-alone: $35.7$, CCD: $16.7$), the CCD prediction is more relevant in both ingredients (broccoli instead of chicken) and cooking instructions (stir instead of sear). 

The video anticipation task remains a challenging task, and both methods fail on steps $5$-$7$ in Figure~\ref{fig:visu_full}(c). 
\section{Conclusion}
We present a novel cross-modal contrastive distillation (CCD) model for instructional activity anticipation. CCD facilities the knowledge distillation between the teacher and student in different modalities. With the help of the contrastive distillation loss, CCD effectively distills the text procedural information from the teacher language model, and transfers this knowledge to help video activity anticipation. Our experiments on the standard Tasty Videos benchmark confirm that CCD improves the anticipation performance over that of the visual-alone baseline by $40.2\%$ relatively in BLEU4, and outperforms the state of the art by a large margin.


\newpage
{
\bibliographystyle{IEEEtran}
\bibliography{egbib}
}

\end{document}